\documentclass[10pt,twocolumn,letterpaper]{article}

\usepackage{wacv}
\usepackage{times}
\usepackage{epsfig}
\usepackage{graphicx}
\usepackage{amsmath}
\usepackage{amssymb}

\usepackage{bm}
\usepackage{multirow}
\usepackage{color}

\usepackage{lipsum}
\newcommand\blfootnote[1]{%
\begingroup
\renewcommand\thefootnote{}\footnote{#1}%
\addtocounter{footnote}{-1}%
\endgroup
}

\newcommand{\tabincell}[2]{\begin{tabular}{@{}#1@{}}#2\end{tabular}}
\def\wacvPaperID{294} % Enter the WACV Paper ID here

\wacvfinalcopy % *** Uncomment this line for the final submission
\ifwacvfinal
\def\assignedStartPage{9876} % *** Enter the assigned starting page number (instead of 9876)
\fi

\ifwacvfinal
\usepackage[breaklinks=true,bookmarks=false]{hyperref}
\else
\usepackage[pagebackref=true,breaklinks=true,colorlinks,bookmarks=false]{hyperref}
\fi

\ifwacvfinal
\else
\fi

\begin{document}

%%%%%%%%% TITLE
\title{JOLO-GCN:  Mining Joint-Centered Light-Weight Information for Skeleton-Based Action Recognition}

% \author{
% % First Author\\
% % Institution1\\
% % Institution1 address\\
% % {\tt\small firstauthor@i1.org}
% % % For a paper whose authors are all at the same institution,
% % % omit the following lines up until the closing ``}''.
% % % Additional authors and addresses can be added with ``\and'',
% % % just like the second author.
% % % To save space, use either the email address or home page, not both
% % \and
% % Second Author\\
% % Institution2\\
% % First line of institution2 address\\
% % {\tt\small secondauthor@i2.org}
% }

\author{
Jinmiao Cai$^{1*}$ \quad Nianjuan Jiang$^{1}$ \quad Xiaoguang Han$^{3}$ \quad Kui Jia$^{2}$ \quad Jiangbo Lu$^{1}$\\

\normalsize $^{1}$SmartMore Corporation\quad 
$^{2}$South China University of Technology\quad 
$^{3}$The Chinese University of Hong Kong (Shenzhen)\\

\tt\small \{jinmiao.cai,nianjuan.jiang,jiangbo\}@smartmore.com hanxiaoguang@cuhk.edu.cn kuijia@scut.edu.cn

% \tt\small \{jinmiao.cai, nianjuan.jiang\}@smartmore.com \quad 
% \tt\small jiangbo.lu@gmail.com \quad 
}

\maketitle
% \thispagestyle{empty}

%%%%%%%%% ABSTRACT
\begin{abstract}
\vspace{-0.05in}
   Skeleton-based action recognition has attracted research attentions in recent years. One common drawback in currently popular skeleton-based human action recognition methods is that the sparse skeleton information alone is not sufficient to fully characterize human motion. This limitation makes several existing methods incapable of correctly classifying action categories which exhibit only subtle motion differences. 
   In this paper, we propose a novel framework for employing human pose skeleton and joint-centered light-weight information jointly in a two-stream graph convolutional network, namely, JOLO-GCN. Specifically, we use Joint-aligned optical Flow Patches~(JFP) to capture the local subtle motion around each joint as the pivotal joint-centered visual information.
   Compared to the pure skeleton-based baseline, this hybrid scheme effectively boosts performance, while keeping the computational and memory overheads low. Experiments on the NTU RGB+D, NTU RGB+D 120, and the Kinetics-Skeleton dataset demonstrate clear accuracy improvements attained by the proposed method over the state-of-the-art skeleton-based methods. \blfootnote{*This work was mainly done when Jinmiao, Nianjuan and Jiangbo were interning and working in Shenzhen Cloudream Technology Co., Ltd. Corresponding email: jiangbo@smartmore.com.}
\end{abstract}

\section{Introduction}
\label{sec:intro}
    Human recognition~\cite{wang2011action,simonyan2014two,wang2016temporal,Feichtenhofer_2019_ICCV} is an active yet challenging task in the field of computer vision. Recently, with the advancement in depth sensors such as Microsoft Kinect and human pose estimation technology~\cite{cao2017realtime, zhou2019hemlets}, obtaining accurate human pose data is becoming much easier. Skeleton-based human action recognition has attracted a lot of attentions and made significant progress over the last decade. Human skeleton motion sequences retain useful high-level motion signal by eliminating the redundant information in the RGB video clips. Compared with the original RGB video clip, a skeleton sequence, with the human body joints in the form of 2D or 3D coordinates, is sparse. Thus, neural networks designed for skeleton-based action recognition can be lightweight and efficient. Recent methods~\cite{yan2018spatial,li2018independently,si2018skeleton,Li_2019_CVPR,shi2019two,Si_2019_CVPR,wang2018beyond,huang2017deep,shi2018non, shi2019skeleton, du2015hierarchical,liu2016spatio,wang2018key} further exploit a variety of deep neural networks in the attempt to fully excavate the internal characteristics of dynamic human skeleton sequences. 
    
    %-----------------------------------------------------------
    \begin{figure}
    	\centering
    	\includegraphics[width=\columnwidth]{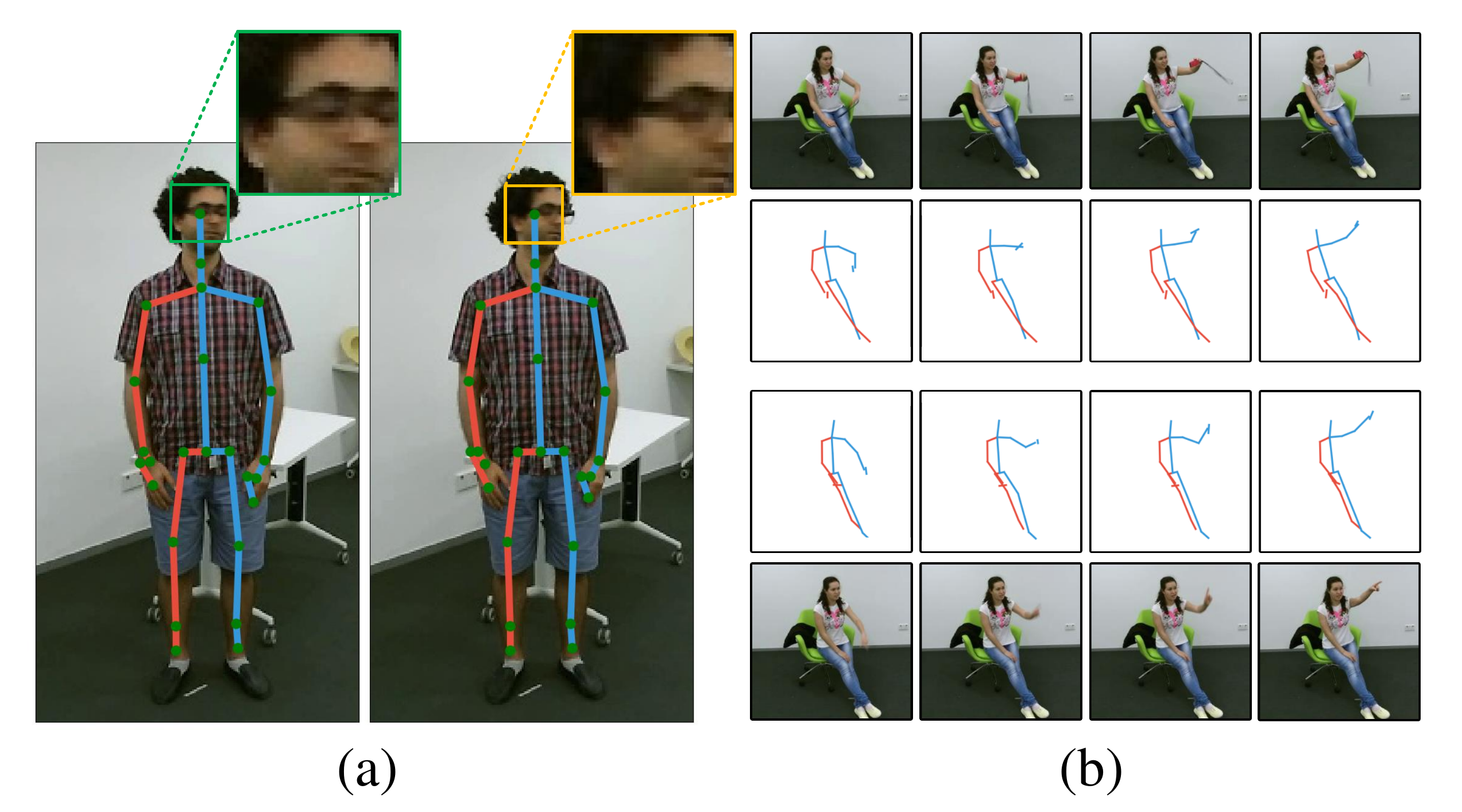}
    	\caption{The motivations of the proposed method. (a)~Sample frames in the ``shake head" sequence from NTU~RGB+D~\cite{Shahroudy_2016_NTURGBD}, where the skeletal joints of the head hardly capture the local subtle movements. (b)~Different actions with similar skeleton sequences. Sample frames are taken from ``taking a selfie" (top row) and ``pointing to something" (bottom row).}
    	\label{fig:motivation}
    	\vspace{-0.1in}
    \end{figure}
    %-----------------------------------------------------------
    
    %=======================================
    % single modal
    As the input of a single-modal action recognition network, skeleton sequences can effectively describe the global human body motion. However, the local subtle motion cues are lost in the process of extracting human poses from video frames. Due to its extreme sparseness, a skeleton sequence can hardly capture subtle features in human motion. There are obvious disadvantages in recognizing human action relying solely on skeleton sequences. 
    
    Firstly, for action categories mainly characterized by local subtle movements, such as ``shaking head"~(Figure~\ref{fig:motivation}(a)), the difference between the skeletons extracted from two successive frames are so subtle that it is hardly useful for describing such actions. The lack of effective representation makes it more challenging for the network to predict related behavioral categories. Moreover, such local subtle movements are easily obscured by the noisy pose estimation, when the body movement of the action is weak. Quantitatively, in regard to the actions such as ``reading" and ``writing", the performance of the skeleton-based single-modal methods seems to have encountered a bottleneck due to the aforementioned drawbacks.
    
    Secondly, skeletons alone may not be as distinctive in describing certain action categories. For instance, as shown in Figure~\ref{fig:motivation}(b), ``pointing to something" and ``taking a selfie" have very similar skeleton sequence representations. Thus these categories are easily confused with each other by a skeleton-based single-modal method. 
    
    % multi modal
    There exist methods that use multi-modal sources to augment skeleton information with some additional inputs, such as RGB features, depth maps, joint heat-maps, for action recognition~\cite{carreira2017quo,luvizon20182d,hu2018deep,baradel2018glimpse,DPIAAAI2019,Liu_2018_CVPR,choutas2018potion,baradel2017human,zolfaghari2017chained,yan2019pa3d}. By introducing such visual information which contains rich motion information both in global and local domains, this kind of multi-modal methods can describe human body motion more completely, and help the neural network to recognize human actions better. However, as the complexity of the input increases, the number of network parameters and the computing resource consumption increase substantially compared to the single-modal counterparts. 
    
    %  solution
    To move beyond such limitations, we propose to augment the 
    skeleton with the light-weight local visual information surrounding skeletal joints in the form of \emph{Joint-aligned optical Flow Patches~(JFP)}. 
    Considering the joint coordinates as the anchors, the most relevant motion cues of the human body can be located in input images. We extract such most relevant motion cues around each human joint from the video as JFP, and keep it light-weight by removing the redundant background information.
    
    After a simple format conversion procedure, a JFP sequence can be represented as sparsely as the skeleton sequence.
    This locally dense but globally sparse representation makes it possible to capture the local subtle motion of human body movement, while keeping the neural network light-weight. 
    The proposed JFP sequence encodes local visual cues with a kinetically meaningful structure inherited from the human pose skeleton.
    Therefore, same as the skeleton sequence, JFPs can be aggregated by very sparse graph connections via a GCN formulation. In this paper, we propose a two-stream GCN architecture, \emph{JOLO-GCN}, to fuse local motion information from a JFP sequence with global motion information from a skeleton sequence. 
    
    To demonstrate the effectiveness of our proposed method, extensive experiments are conducted on three popular large-scale human action recognition datasets, namely, NTU-RGB+D~\cite{Shahroudy_2016_NTURGBD}, NTU RGB+D 120~\cite{Liu_2019_NTURGBD120} and Kinetics-Skeleton~\cite{carreira2017quo,yan2018spatial}. Our method outperforms state-of-the-art skeleton-based methods on these datasets. The main contributions in this paper are summarized as follows:
    \begin{itemize}
    	\item We propose a novel scheme to represent the visual information surrounding each skeletal joint as JFP, which contains rich local subtle body motion cues in a relatively sparse format.
    	\vspace{-0.1in}
        \item We demonstrate that our JOLO-GCN which jointly uses the local motion information from a JFP sequence with the global motion information from a skeleton sequence, gains significant accuracy improvements compared with the single-modal baselines.
    	\vspace{-0.1in}
    	\item Extensive experiments are conducted on three large-scale human action recognition datasets. Our method obtains the state-of-the-art results on these datasets.
    \end{itemize}

\section{Related Work}
    In this section, we briefly review the approaches that have utilized human skeletons as a key information source for action recognition. These approaches can be divided roughly into single-modal skeleton-based methods and multi-modal skeleton-based methods.
    
    \subsection{Single-Modal Skeleton-Based Methods}
    Taking only the skeleton data as the input for the human action recognition task, several single-modal methods~\cite{yan2018spatial,li2018independently,si2018skeleton,Li_2019_CVPR,shi2019two,Si_2019_CVPR,wang2018beyond,huang2017deep,shi2018non, shi2019skeleton,du2015hierarchical,liu2016spatio} have been proposed. These methods are usually light-weight and computationally efficient. 
    
    % GCN spatial
    Graph convolutional networks (GCN) use a general and effective framework for learning representation of graph structured data. Various GCN variants~\cite{shi2019skeleton,yan2018spatial,shi2019two,Li_2019_CVPR,Si_2019_CVPR} have achieved the state-of-the-art results on skeleton-based action recognition. Yan~\etal proposed ST-GCN~\cite{yan2018spatial}, a generic graph-based formulation for modeling dynamic skeletons, which can capture the patterns embedded in the spatial configuration of the joints as well as their temporal dynamics. 
    The topology of the graph employed in ST-GCN is fixed and defined according to the human skeletal structure. Thus, it is not guaranteed to be optimal flexibly for recognizing specific actions. In the attempt to address this drawback, adaptive graph CNNs~\cite{shi2019two,Li_2019_CVPR,shi2019skeleton}, which can adaptively learn the graph topology, are used for automatically inferring spatial dependency among pose joints. 
    In AS-GCN~\cite{Li_2019_CVPR}, the actional links~(A-link) is inferred in a data-driven manner to capture the latent dependencies between any pose joints.
    In 2S-AGCN~\cite{shi2019two}, three parts -- a fixed physical structure graph, a data-dependent attention graph and a global learned graph -- are designed in the adjacency matrix of an adaptive graph, which increase the model flexibility and also the generality. In addition, the second-order information of joint coordinates (namely "Bones")
    %, which represents the feature of bones between two joints, 
    is aggregated together with the ``Joints" as a parallel source input for providing more motion cues.
    
    % RNN LSTN temporal
    The RNN and LSTM structures are effective for analyzing time series of streaming data. Different methods based on a RNN or LSTM framework are explored in recent works~\cite{du2015hierarchical,li2018independently, Si_2019_CVPR,si2018skeleton,wang2018beyond,liu2016spatio}. 
    In IndRNN~\cite{li2018independently}, the neurons in the same layer are independent of each other, which has great stability against gradient vanishing and exploding. 
    Si~\etal \cite{si2018skeleton} proposed SR-TSL, which utilizes a LSTM module in series with GCN for spatial reasoning and temporal stack learning. AGC-LSTM~\cite{Si_2019_CVPR} applies a graph convolution operator to replace the fully connected operator within LSTM, so as to explore the co-occurrence relationship between spatial and temporal domains. %Liu~\etal \cite{liu2016spatio} extended LSTM to the spatial-temporal domain to explicitly model the dependencies between joints and introduced a new gating mechanism to handle noise and occlusion in the skeleton data.
    
    Since the skeleton input is sparse, these single-modal skeleton-based methods are advantageous in terms of computational complexity and the network scale. However, subtle motion characteristics that cannot be well captured by skeleton dynamics remain challenging for single-modal skeleton-based networks to learn.
    
    \subsection{Multi-Modal Skeleton-Based Methods}
    Multi-modal approaches are widely used in the field of action recognition~\cite{carreira2017quo,luvizon20182d,hu2018deep,baradel2018glimpse,DPIAAAI2019,Liu_2018_CVPR,choutas2018potion,baradel2017human,zolfaghari2017chained,yan2019pa3d}. Augmenting data sources, such as RGB images, optical flow, depth maps, joint heat-maps, and pose skeletons, provide richer semantic cues for neural networks to infer the human action in the scene. Since a comprehensive literature review is beyond the scope of this paper, here we only focus on skeleton-based multi-modal methods.
    
    The chained multi-stream network~\cite{zolfaghari2017chained} computes and integrates several important visual cues 
    % -- human body part segmentation, optical flow, and the RGB image 
    in a Markov chain model which adds complementary cues successively. The C3D architecture~\cite{tran2015learning} is used as the base architecture of the multi-stream network.
    Hu~\etal~\cite{hu2018deep} proposed a novel deep bilinear framework for learning multi-modal temporal features. Multi-modal inputs include RGB images, depth maps, and skeleton sequences. Note that the depth and RGB input utilized in this work~\cite{hu2018deep} are image patches aligned with the skeletal joints. 
    % However, the spatial independence among different patches is tampered after tiling the local patches to compose an integrated image and the subsequent convolution operation.
    Luvizon~\etal \cite{luvizon20182d} proposed a multi-task deep architecture to perform 2D and 3D pose estimation jointly with action recognition. The idea is to aggregate image visual features, joint probability maps and the estimated pose sequence to infer the human action in the recognition module.
    Joint heat-maps are utilized in~\cite{DPIAAAI2019,Liu_2018_CVPR,choutas2018potion} to augment the sparse skeleton information. High-dimensional heat-map sequences are compressed in different representations. Additional motion cues might be inferred from such probabilistic representations for recognizing action.

	Despite the benefits of bringing more cues to facilitate the action recognition task, these multi-modal approaches can only handle small temporal windows due to the significant amount of network parameters and computing resources required. 
	% Moreover, these complex methods have not yet shown large accuracy improvements over baseline methods.

	%--------------------------------------------
	\begin{figure}
		\centering
		\includegraphics[width=8.2cm,height=7.1cm]{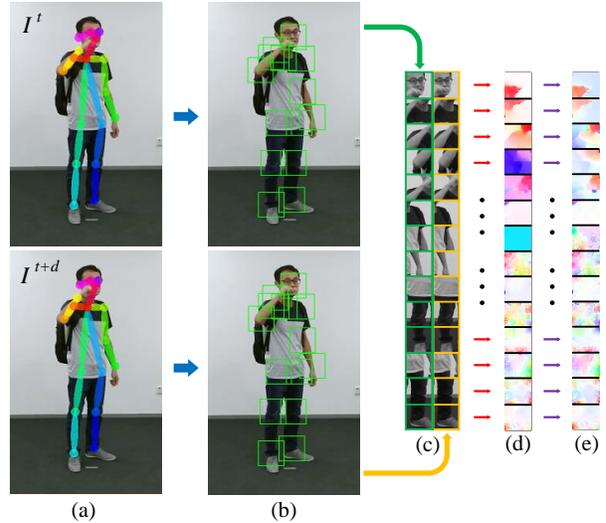}
		\caption{The Joint-aligned optical Flow Patches~(JFP) estimation from two successive frames with corresponding 2D pose joints. (b) Patches centered around body joints obtained for (a) a successive frame pair are first cropped to form (c) JAP pairs. (d) Residual optical flow is estimated from each JAP pair, followed by zero-mean normalization to obtain the corresponding (e) JFP. Noted that the zero-mean normalization is applied for eliminating the influence of joint localization imprecision on the joint alignment.}
		\vspace{-0.07in}
		\label{fig:JAPJFP}
	\end{figure}
	%--------------------------------------------

\section{Method}
\vspace{-0.02in}
	Given a video sequence, we denote the image frames as $\mathcal{I} = \{ I^t\in\mathbb{R}^{H\times W\times 3} $~$\mid$~$t=1,...,T \}$, where $H$ is the height of the video frame, and $W$ denotes the frame width. The corresponding skeleton sequence 
	containing the 2D/3D coordinates of \emph{K} joints in all \emph{T} frames, is denoted as $\mathcal{J} = \{ J_k^t$~$\mid$~$k=1,...,K; t=1,...,T \} $.
	We first use JFP to augment the 2D/3D pose skeleton sequence, and then use GCN for feature embedding and learning.
	
	\subsection{Joint-Aligned Optical Flow Patches~(JFP)}
	\label{subsec:JFP}
% 	Conventional skeleton sequences characterize the human body motion by sparse 2D/3D coordinates. Though they effectively describe the human body motion in a global domain, a skeleton sequence lacks the ability to capture vital visual cues for subtle human motion, which are actually crucial for recognizing certain action categories. 
	
	As analyzed in the introduction, the sparse skeleton information alone is not sufficient to fully characterize human motion.
	In the attempt to address this limitation, we treat the human joint coordinates as anchor points in the image, and explicitly describe subtle motion cues corresponding to these locations in the form of optical flow patches. 
	
	Specifically, a local square-shaped patch $W_k^t$ is obtained for each joint $J_k^t$ in frame $I^t$ by a \emph{joint-centered}~(w.r.t. the 2D coordinates of the joint) cropping operation:
	\vspace{-0.05in}
	\begin{equation}
	\label{eq:crop}
    	W_k^t = \bm{Crop}(I^t, J_k^t,~l)\;,
	\end{equation}

	\noindent where $\bm{Crop}(\cdot)$ denotes the joint-centered cropping operation, and $l$ is a custom parameter that denotes the side length of the cropped patch $W_k^t$. We term $W_k^t$ as Joint-centered Appearance Patch~(JAP), and the 
	corresponding JAP sequence for the skeleton sequence $\mathcal{J}$ is denoted by $\mathcal{W} = \{W_k^t$~$\mid$~$k=1,...,K; t=1,...,T \}$. 
	
	It is clear that the appearance of body parts can be explicitly captured in JAPs. 
	In addition, a JAP sequence also presents the local subtle motion of each body part.
	% In fact, for the temporal domain, a JAP sequence also presents itself as a viable source to extract the local subtle motion of body parts.
	Intuitively, optical flow explicitly reflects the dense motion field between successive video frames, and is a more effective representation for subtle motion cues. Motivated by this, we propose to estimate the optical flow between successive JAP pairs of each body joint to obtain the Joint-aligned optical Flow Patch~(JFP)~(see Figure~\ref{fig:JAPJFP}). The JFP sequence corresponding to $\mathcal{J}$ is denoted as $\mathcal{F} = \{F_k^t$~$\mid$~$k=1,...,K; t=1,...,T \}$.

	% To eliminate the influence of joint localization imprecision on the estimated optical flow, we further normalize the flow magnitude of each JFP individually using zero-mean normalization. This also magnifies the characteristics of different local motion cues despite their varying signal strength levels. 
	
\begin{figure}
		\centering
		\includegraphics[width=\columnwidth]{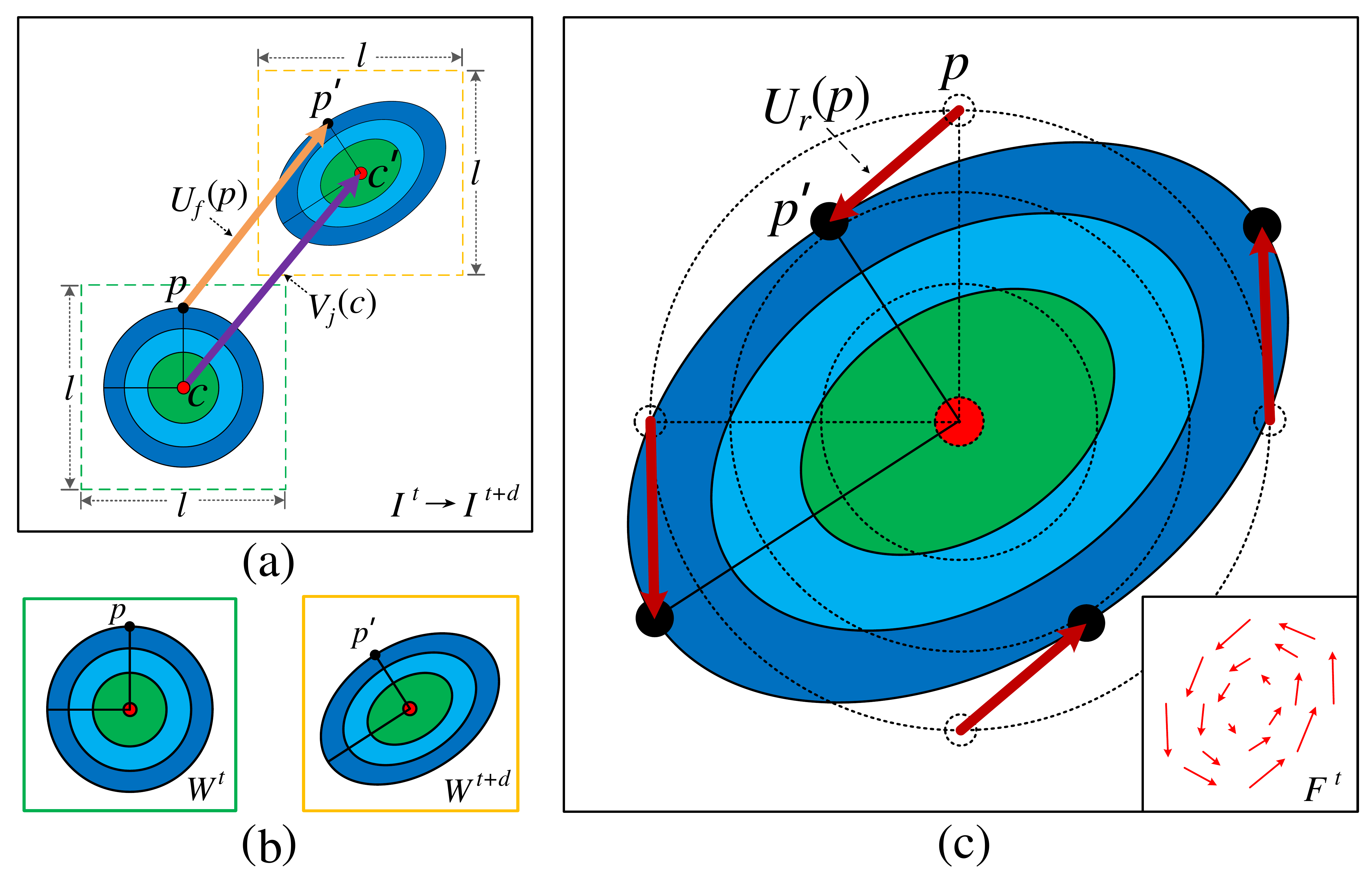}
		\caption{Schematic diagram of Joint-aligned optical Flow Patch (JFP) for two successive frames. (a)~The object motion \bm{$U_f$} from frame $I^t$ to frame $I^{t+d}$, including translation, rotation and deformation. (b)~The corresponding JAPs: $W^t$ and $W^{t+d}$. (c)~The alignment of two JAPs and the local subtle motion field {$U_r$}, which is computed to generate JFP $F^t$.}
		\label{fig:JFP_motion_field}
	\end{figure}

	Unlike the conventional dense optical flow, 
	JFP only focuses on capturing the local subtle motion. For a clear description of the local motion field captured in JFP, we decompose the general motion field between consecutive frames as follows.
	As illustrated in Figure~\ref{fig:JFP_motion_field}(a), an object (or a body part concerned in this work) moves from bottom-left to top-right between two successive video frames $I^t$ and $I^{t+d}$, where $d$ denotes a temporal interval. In this process, the object may undergo translation, rotation and deformation. The motion vector {$U_f$}$(p)$ of the matching pixels $p$ and $p'$ can be obtained by estimating the conventional optical flow between these  two full frames. In fact, the global displacement of the object can be represented simply as the movement of the object center, i.e., 
	a motion vector {$V_j$}$(c)$ from the object center location $c$ to its counterpart $c'$.
	
	After center-aligning the JAP pairs $W^t$ and $W^{t+d}$ in Figure~\ref{fig:JFP_motion_field}(b), the global displacement is actually eliminated. When computing optical flow from the JAP pair, the resulting residual motion field {$U_r$} in the JFP $F^t$ as shown in Figure~\ref{fig:JFP_motion_field}(c) only contains subtle local motion such as rotation and deformation. 	Based on the local motion field {$U_r$} and the global displacement {$V_j$}$(c)$ by differencing the corresponding joints' coordinates, the full motion {$U_f$}$(p)$ at pixel $p$ can be approximated as follows:
	\vspace{-0.05in}
	\begin{equation}
    \label{eq:field}
    	{U_f}(p) \approx {V_j}(c) + {U_r}(p) \;. 
	\end{equation}
	Around an object (or a body part concerned in this work), Eqn.~(\ref{eq:field}) gives an approximated expression on the relationship of the residual local motion field ${U_r}$ captured in JFP with the full motion field ${U_f}$ in the conventional optical flow and the global displacement vector {$V_j$}$(c)$.
	Please note that JFP is estimated from the JAP pair directly, but not computed as the approximated expression in Eqn.~(\ref{eq:field}).
	The residual local motion field ${U_r}$ captured in JFP is the key information used in our method for action recognition.
	
	We observe that the proposed JFP representation has two nice properties. Firstly, JFP provides only local subtle motion cues. As discussed above, the motion of human body parts can be represented as the global joint displacement plus the local subtle motion. The information captured by JFP is orthogonal to that of skeleton coordinates, which reflect the global joint displacement.
	% The sparse joint coordinates in a skeleton sequence only provide cues for the global joint displacement and are unable to reflect local subtle motions. 
	%The information captured by JFP is orthogonal to that of skeleton coordinates. 
	Secondly, compared with general dense visual data, a JFP sequence encodes visual cues with a kinetically meaningful structure inherited from the human pose skeleton, namely, JFPs have a one-to-one correspondence with multiple skeletal joints. Although there are alternative visual features with a similar property, such as human pose heat-maps. 
	% and part segmentation probability maps, 
	Their data resolution is relatively large or they need to be learned by a neural network with substantially more parameters. 
	% Also, they do not explicitly capture the local motion information across frames.

	%----------------------------------------------
	%---------------------------------------------

    \subsection{Graph Convolutional Networks~(GCN)}
    \label{subsec:GCN}
	In order to better capture the kinetically structured property encoded in the JFP sequences, we adopt GCN as our backbone network architecture. 
	Given a $K$ joints skeletal graph with a node collection of $\mathcal{V}$ = $\{v_{k}$~$\mid$~$v_{k}=J_k, k=1,...,K\}$,
% 	Given a skeleton with $K$ joints, the graph is constructed as $\mathcal{G} = \{\mathcal{V}, \mathcal{E}\}$, where $\mathcal{V}$ = $\{v_{k}$~$\mid$~$v_{k}=J_k, k=1,...,K\}$ denotes the set of graph nodes corresponding to the joints and $\mathcal{E}$ denotes the set of graph edges corresponding to the skeletal connection between joints.
	the neighborhood of a node $v_{i}$ is defined as $\mathcal{N}$($v_{i}$) = $\{v_j$~$\mid$~$d(v_{i}, v_{j}) \le D  \}$, where $d(v_{i}, v_{j})$ is the shortest path length from $v_{j}$ to $v_{i}$.
	In a ST-GCN network setup~\cite{yan2018spatial}, the spatial graph convolution unit is the key component, which is constructed to capture the spatial feature among joints. More specifically, given the graph adjacency matrix $\bm{A} \in \mathbb{R}^{K\times K}$, $\bm{A}(j,i)=w$ if $v_j \in \mathcal{N}(v_i)$ and $\bm{A}(j,i)=0$ otherwise. The adjacency matrix is normalized using a degree matrix $\bm{\varLambda}$ as:
    \vspace{-0.05in}
	\begin{equation}
	\label{eq:gcn-matrix}
	\begin{array}{lr}
	\bm{\varLambda}(i,i) = \sum\limits_{j=1}^K \bm{A}(i,j); \;\; \bm{A}^{\bm{norm}}=\bm{\varLambda}^{-\frac{1}{2}}\bm{A}\bm{\varLambda}^{-\frac{1}{2}}\;.
	\end{array}
	\end{equation}
	
	For the spatial graph convolution can be written in terms of the adjacency matrix as follows:
	\begin{equation}
	\label{eq:gcn-stgcn}
	\begin{array}{lr}
	\bm{Y}_{out} = \sigma(\bm{W}(\bm{X}_{in}(\bm{A}^{norm}\odot\bm{M}))) \;,
	\end{array}
	\end{equation}
	
	%----------------------------------------------
	\begin{figure*}
		\centering
		\includegraphics[width=15cm,height=4.7cm]{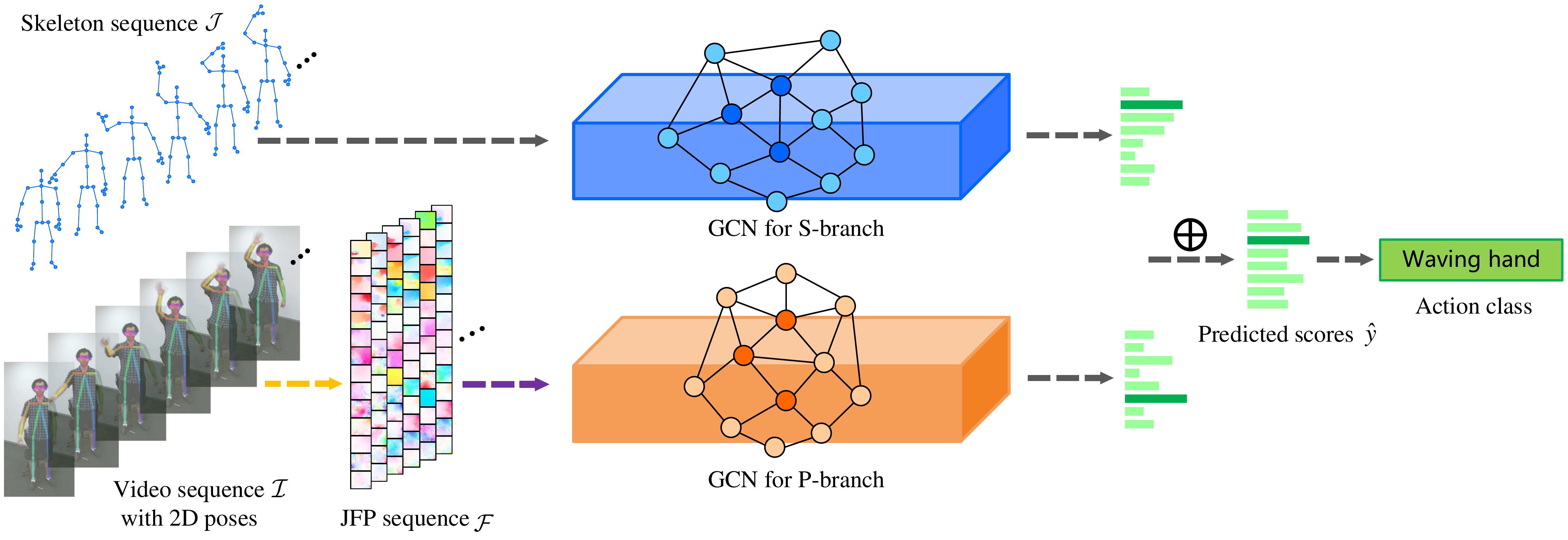}
		\caption{Framework of the proposed JOLO-GCN. A two-stream GCN-based network architecture is used to process the skeleton sequence (S-branch) and the JFP sequence (P-branch) independently. 
		The predicted scores from these two branches are merged into the final predicted scores for each action class by linear blending.
		}
		\label{fig:framework}
		\vspace{-0.1in}
	\end{figure*}
	%----------------------------------------------
	
	\noindent where $\bm{Y}_{out} \in \mathbb{R}^{d_{out}\times K}$, $\bm{X}_{in} \in \mathbb{R}^{d_{in}\times K}$ and $d$ denotes the feature dimension. $\bm{M}$ denotes a trainable mask for the adaptive re-weighting on the $\bm{A}^{norm}$, and $\odot$ denotes the Hadamard product. The operation of $(\bm{X}_{in}(\bm{A}^{norm}\odot\bm{M}))$ guarantees the features corresponding to different skeletal joints to interact by following the skeleton topology given by $\bm{A}$. $\bm{W}$ denotes a $1\times1$ convolution layer to expand the feature dimension and $\sigma(\cdot)$ is a non-linear activation layer (e.g., ReLU).
	% For the temporal dimension, a temporal convolution unit is subsequently connected to the spatial graph convolution unit. For each graph node, it performs a $K_t\times1$ convolution over the features of the corresponding joints in a $K_t$-sized temporal window. 
	We refer interested readers to \cite{yan2018spatial,niepert2016learning} for a detailed discussion on GCNs for skeleton-based action recognition and GCNs in the context of the spatial graph theory.

	\subsection{JOLO-GCN}
	\label{subsec:Framework}
	The framework of JOLO-GCN is illustrated in Figure~\ref{fig:framework}. We use a two-stream network architecture for processing the skeleton sequence and the JFP sequence independently. 
	% To better exploit the implicit kinetic structure property, we adopt GCN as the backbone of the two-stream network.
	%As shown in Figure~\ref{fig:framework}, 
	The skeleton sequence $\mathcal{J}$ is fed into the first GCN branch~(called {\it S-branch}) and the prediction scores are generated subsequently. In parallel, the precomputed JFP sequence $\mathcal{F}$ is fed into the second GCN branch~(called {\it P-branch}). We train S-branch and P-branch independently. The cross-entropy loss $\mathcal{L}$ is used for the training:
	\begin{equation}
	\label{eq:loss}
    \begin{array}{lr}
        \mathcal{L} =-y^{T}log(\hat{y})
	\end{array}
	\end{equation}
	\noindent where $y$ denotes a one-hot label vector of the ground-truth action class and $\hat{y}$ denotes the predicted scores. The final action class prediction scores are obtained by linear-blending the predicted scores from the two GCN branches.

	\subsection{Data Format Conversion of JFP}
	\label{subsec:conversion}
	A RGB video sequence $\mathcal{I}$ takes on a four-dimensional format of \emph{T}$\times$\emph{H}$\times$\emph{W}$\times$3. Compared with the original video sequence, the skeleton sequence $\mathcal{J}$ is in a smaller four-dimensional format of \emph{T}$\times$\emph{K}$\times$3$\times$\emph{N}, with the dimension \emph{N} representing the maximum number of persons that may appear in the scene. The magnitude of \emph{K}$\times$\emph{N} is much smaller than \emph{H}$\times$\emph{W}. As a consequence, though the scale of the neural network for a video sequence would be large usually, the skeleton sequence can be processed with some lightweight networks, such as RNNs or GCNs.
	Based on the earlier discussion, the proposed JFP sequence is considered as an orthogonal cue to the skeleton sequence. In terms of the data format, the JFP sequence is designed to hold the sparsity property as the skeleton sequence. This assures that it can be processed in a lightweight network accompanied with the skeleton sequence. 
	% In order to achieve this goal
	Therefore, we downsample each JFP using a bilinear interpolation function from the resolution of $l$$\times$$l$ to a smaller one of $\mu$$\times$$\mu$. Then, we convert the data format of a JFP sequence similar to that of the skeleton sequence. As the format of the skeleton sequence is \emph{T}$\times$\emph{K}$\times$3$\times$\emph{N}, the JFP sequence~(\emph{T}$\times$\emph{K}$\times$$\mu$$\times$$\mu$$\times$2$\times$\emph{N}) is converted to a four-dimensional format of \emph{2T}$\times$\emph{K}$\times$$\mu^2$$\times$\emph{N}. This design choice keeps a good trade-off between the strength of the JFP representation and the computational complexity.

    % =======================================
	\begin{table*}[t!]
		\setlength{\abovecaptionskip}{-1pt}
		\centering
		\begin{center}
			\begin{tabular}{|c|c|c|c|c|c|c|c|c|c|}
				\hline
				\multicolumn{2}{|c|}{Method}  
				& Pose & Visual 
				& \tabincell{c}{NTU \\ X-sub~(\%)} &\tabincell{c}{NTU \\ X-view~(\%)} 
				& \tabincell{c}{NTU 120\\X-sub~(\%)} &\tabincell{c}{NTU 120\\X-setup~(\%)} 
				& \tabincell{c}{KS \\ Top-1~(\%)}   &  \tabincell{c}{KS \\ Top-5~(\%)} \\ 
				\hline
				
				\multicolumn{2}{|r|}{GCA-LSTM\cite{liu2017skeleton}} 
				& \checkmark & - 
				& 76.1 & 84.0
				& 61.2 & 63.3
				& - & -\\

				\multicolumn{2}{|r|}{SkeleMotion\cite{caetano2019skelemotion}} 
				& \checkmark & -
				& 76.5 & 84.7
				& 67.7 & 66.9
				& - & -\\
				
				\multicolumn{2}{|r|}{Chained Net\cite{zolfaghari2017chained}} 
				& \checkmark & \checkmark 
				& 80.8 & - 
				& - & - 
				& - & -\\
				
				\multicolumn{2}{|r|}{ST-GCN\cite{yan2018spatial}} 
				& \checkmark & - 		   	
				& 81.5   & 88.3 
				& - & -
				& 30.7   & 52.8  \\
				
				\multicolumn{2}{|r|}{Ind-RNN\cite{li2018independently}}	 
				& \checkmark & -
				& 81.8   & 88.0
				& - & -  
				& - & -\\
				
				\multicolumn{2}{|r|}{SR-TSL\cite{si2018skeleton}}	
				& \checkmark & - 	
				& 84.8   & 92.4   
				& - & - 
				& - & -\\	
				
				\multicolumn{2}{|r|}{Deep Bilinear\cite{hu2018deep}}  
				& \checkmark & \checkmark 	
				& 85.4  & 90.7   
				& - & - 
				& - & -\\
				
				\multicolumn{2}{|r|}{2D/3D pose\cite{luvizon20182d}}  
				& \checkmark & \checkmark 	
				& 85.5  & -   
				& - & - 
				& - & -\\
				
				\multicolumn{2}{|r|}{AS-GCN\cite{Li_2019_CVPR}}     
				& \checkmark & - 		   
				& 86.8   & 94.2   
				& - & -
				& 34.8   & 56.5 \\
				
				\multicolumn{2}{|r|}{2S-AGCN\cite{shi2019two}}	
				& \checkmark & - 	
				& 88.5   & 95.1   
				& 83.7* & 85.8*
				& 36.1   & 58.7  \\

				\multicolumn{2}{|r|}{SGN\cite{Zhang_2020_CVPR}}	
				& \checkmark & - 	
				& 89.0   & 94.5 
				& 79.2   & 81.5
				& -   & -  \\
				
				\multicolumn{2}{|r|}{AGC-LSTM\cite{Si_2019_CVPR} }  
				& \checkmark & - 	
				& 89.2   & 95.0   
				& - & - 
				& - & -\\
				
				\multicolumn{2}{|r|}{MS-G3D\cite{Liu_2020_CVPR}} 
				& \checkmark & - 
				& 91.5  & 96.2
				& 86.9 & 88.4
				& 38.0 & 60.9 \\
				
				\multicolumn{2}{|r|}{Posemaps\cite{Liu_2018_CVPR}} 	 	
				& \checkmark & \checkmark 
				& 91.7  & 95.3  
				& 64.6 & 66.9
				& - & -\\
				
				\hline
				\hline
				& Backbone               		
				& Pose       & Visual 
				& X-sub(\%) & X-view(\%) 
				& X-sub(\%) & X-setup(\%)
				& Top-1(\%) & Top-5(\%) \\
				
				\hline
				\multirow{2}{*}{\textbf{Ours}} & ST-GCN 
				& \checkmark & \checkmark 
				& 90.4 & 95.6 
				& - & -
				& 33.7 & 57.6 \\
				
				& 2S-AGCN & 
				\checkmark & \checkmark 			
				& \textbf{93.8}& \textbf{98.1} 
				& \textbf{87.6} & \textbf{89.7}
				& \textbf{38.3}   & \textbf{62.3}  \\
				\hline
			\end{tabular}
		\end{center}
		\caption{Quantitative comparisons of the validation accuracy on the NTU RGB+D, NTU RGB+D 120 and Kinetics-Skeleton datasets~(referred to as ``NTU", ``NTU 120", and ``KS", respectively). 
		The proposed JOLO-GCN is built and evaluated over two different GCN-based backbones~(ST-GCN~\cite{yan2018spatial} and 2S-AGCN~\cite{shi2019two}), and we report both end results accordingly. 
		The second and third columns~(``Pose" and ``Visual") indicate whether a method under evaluation utilizes skeleton and/or visual data~(e.g. images, depth maps, and optical flow), respectively. * denotes the results obtained using the released codes.}
		\label{tab:NTU+KS}
	\end{table*}
	% =======================================

\section{Datasets and Implementation Details}
\label{sec:Dateset_and_Implementation_Details}
	\subsection{Datasets}
	We evaluate the performance of our method on three benchmark skeleton-based action recognition datasets.
	
	\textbf{NTU RGB+D:} ``NTU RGB+D"~\cite{Shahroudy_2016_NTURGBD} is a widely-used benchmark dataset in the field of skeleton-based human action recognition. In this dataset, 56,880 video samples corresponding to 60 action classes are provided. All samples are performed by 40 distinct performers and recorded in 17 different indoor scene setups by three cameras from different views. The provided data of each sample include an RGB video, a 3D human skeleton sequence, a depth map sequence and a IR video. Two official evaluation protocols are adopted in our experiments, i.e., Cross-subject~(X-sub) and Cross-view~(X-view).

	\textbf{NTU RGB+D 120:}
	``NTU RGB+D 120" is the extended version of the NTU RGB+D dataset by adding another 60 more challenging classes with another 57,600 video samples. All video samples are performed by 106 distinct performers in a wide range of age distribution, and recorded in 32 different indoor scene setups by three cameras from different views. Two evaluation protocols are recommended: Cross-subject~(X-sub) and Cross-setup~(X-setup).

	\textbf{Kinetics-Skeleton:} Kinetics~\cite{carreira2017quo} is a large-scale dataset for action recognition. It contains around 300,000 video clips. 
	%All of the clips have gone through multiple rounds of human annotations, and each is taken from a unique YouTube video. 
	The actions cover 400 classes ranging from daily activities, sports scenes, to complex actions with interactions.
	% Kinetics provides only raw video clips but without skeleton data. 
	A related dataset named Kinetics-Skeleton~\cite{yan2018spatial} is generated for skeleton-based action recognition. Adopting the public OpenPose toolbox~\cite{cao2017realtime}, the authors estimated the 2D pose of 18 joints for every frame of the video clips, and also attained each joint's estimation confidence. 
	% Then, based on the average joint confidence, two people are selected for those multi-person clips. 
	Following the evaluation method in~\cite{yan2018spatial,shi2019two,shi2019skeleton,Li_2019_CVPR}, the dataset is divided into a training set (240,000 samples) and a validation set (20,000 samples). Top-1 and Top-5 accuracies are reported.
	%We use the Top-1 and Top-5 accuracy on the validation set as the evaluation protocols.

	\subsection{Implementation Details}
	\textbf{Implementing JFP:} In order to improve experimental efficiency, the JFP sequences are estimated from video sequences and 2D pose sequences in advance.
	%before the network training and evaluation. 
	Following the setting of \cite{DPIAAAI2019,Liu_2018_CVPR}, we use the OpenPose toolbox~\cite{cao2017realtime} to extract the corresponding 2D pose sequence~(i.e., 18 joints) from videos. We remove the joints of eyes and ears to eliminate the redundancy of the patch overlap between adjacent joints. As a result, a total of 14 joints are chosen for processing the corresponding JFPs. In this work, we use the classic TV-L1 algorithm~\cite{zach2007duality} for the optical flow estimation due to its simplicity and effectiveness.
	
	For the NTU RGB+D and NTU RGB+D 120 dataset, the scale of human bodies in the scene is relatively stable, so the patch size $l\times l$ of JFP is empirically set to a fixed value i.e., $l=32$. For the Kinetics-Skeleton dataset, because the scale of the human body has a large variation, we use the average bone length of each sample to define the patch size. 
	% Specifically, we connect the related joint pairs to form bone vectors for the whole human skeleton, and calculate the average length of bone vectors for the entire video sequence.
	Finally, all of these JFPs are downsampled to a smaller resolution of $\mu\times \mu$, with $\mu = 8$ in our experiment.

	\textbf{Two-Stream GCN-Based Network:} In order to fully verify the effectiveness and general compatibility of the JOLO-GCN, we opt for two different GCN-based backbones in our experiments: 1) ST-GCN~\cite{yan2018spatial}, and 2) 2S-AGCN~\cite{shi2019two}. In our algorithm design, the joint number of the JFP sequences is 14, while the skeleton sequences use a different 25-joint skeleton structure. Therefore, different graph structures following the respective skeleton structures are constructed for the S-branch and the P-branch.
	
	In the S-branch, an input skeleton sequence is required to contain 300 frames in the original settings of ST-GCN and 2S-AGCN.
	%, but this frame number requirement actually exceeds the average length of all video samples. As a result, when the length of a given video is shorter than 300 frames, empty joint coordinates with their values set to $0$ are added at the end to make up for 300 frames in total. This design choice affects the learning system's classification performance by unnecessarily injecting dummy padding frames. In view of this, when using the ST-GCN backbone, we shorten the time window length of the skeleton sequences from 300 to 128, i.e., \emph{T=}128. As for 2S-AGCN~\cite{shi2019two}, we follow the original setting of repeating the skeleton sequence until filling up the 300-frame time window. 
	In the P-branch, the input JFP sequences are downsampled by a temporal downsampling factor of $2$, and the sequence length of the corresponding JFP is 64. 
	% This allows the skeleton sequence and the JFP sequence to cover the same length of a given video sequence. The JFP sequences are hence repeated until they reach 64 frames in length. Finally, 
	The JFP sequence~(\emph{T}$\times$\emph{K}$\times$$\mu$$\times$$\mu$$\times$2$\times$\emph{N} = 64$\times$14$\times$8$\times$8$\times$2$\times$2) is converted to \emph{2T}$\times$\emph{K}$\times $$\mu^2$$\times$\emph{N} = 128$\times$14$\times$64$\times$2.

	The predicted scores from the two branches are added with blending weights to obtain the final prediction. The scores from the S-branch and the P-branch are merged by linear-blending with the weights of 0.5 and 0.5. 

\section{Experiments and Analysis}
\label{sec:Experiment}
	\subsection{Experimental Results}
	\vspace{-0.05in}
	We evaluate the proposed JOLO-GCN on the NTU RGB+D, NTU RGB+D 120, and Kinetics-Skeleton datasets and compare our method with the state-of-the-art skeleton-based action recognition methods. 
	% The comparison results are shown in Table~\ref{tab:NTU+KS}.
	As shown in Table~\ref{tab:NTU+KS}, our best model based on the 2S-AGCN backbone obtains action classification accuracy of 93.8\% and 98.1\% for the X-sub and X-view protocols on NTU RGB+D, 87.6\% and 89.7\% of classification accuracy for X-sub and X-setup protocols on NTU RGB+D 120 dataset, and 38.3\% and 62.3\% of Top-1 and Top-5 accuracy on Kinetics-Skeleton. These results validate the new state-of-the-art accuracy achieved by the proposed JOLO-GCN on the three benchmark datasets. In addition, when compared with the original GCN-based baselines, i.e., ST-GCN~\cite{yan2018spatial} and 2S-AGCN~\cite{shi2019two}, our proposed method integrated with the JFP stream significantly improves their accuracy by 8.9\% and 5.3\% on the NTU RGB+D X-sub protocol, respectively.

	It is clear to see that our method outperforms all the single-modal methods~\cite{huang2017deep,soo2017interpretable,yan2018spatial,li2018independently,si2018skeleton,ijcai2018-109,Li_2019_CVPR,shi2019two,Si_2019_CVPR,shi2019skeleton} owing to the introduced JFP stream, which provides useful local motion information. Meanwhile, the proposed JOLO-GCN also outperforms all the multi-modal methods, which utilize skeleton and/or visual data, such as human pose heat-map~\cite{DPIAAAI2019,Liu_2018_CVPR}, depth maps~\cite{hu2018deep}, and optical flow~\cite{zolfaghari2017chained}.

	\subsection{Ablation Study}
	\vspace{-0.05in}
	In this subsection, we examine the proposed JFP stream in more depth, and in particular the performance improvement brought by JFP and its advantage over other possible modalities for action recognition. %Since the X-sub protocol on the NTU RGB+D dataset is more challenging than the X-view protocol, our ablation studies are mainly carried out on this metric. The experiments are conducted over both the ST-GCN and the 2S-AGCN backbones, to validate the proposed method's general applicability and accuracy improvements.
	%-------------------------------------------------------
	\begin{table}[t]
		\centering
		\begin{center}
			\begin{tabular}{|c|l|c|}
				\hline
				Backbone & Data used  & X-sub~(\%) \\
				\hline
				& Joints~(S-branch)      			 		  	& 81.5  	   \\
				ST-GCN	& JFP~(P-branch)     		 			& 86.6 		   \\
				& Joints + JFP    		        				&\textbf{90.4} \\
				\hline
				& Joints  				  						& 86.6         \\
				& Bones    				  						& 86.2         \\
				2S-AGCN& Joints + bones~(S-branch)   		  	& 88.5         \\
				& JFP~(P-branch) 			 					& 88.1         \\
				& Joints + bones + JFP 	           				&\textbf{93.8} \\
				\hline
			\end{tabular}
		\end{center}
		\caption{Comparisons of the accuracy obtained by the S-branch, the P-branch and their combination on the NTU RGB+D dataset.}
		\label{tab:comparison_backbones_inputs}
		\vspace{-0.1in}
	\end{table}
	%-------------------------------------------------------

	\textbf{Comparison between S-branch and P-branch:}
	As shown in Table~\ref{tab:comparison_backbones_inputs}, when ST-GCN is used as the GCN backbone,
    the result of the P-branch using the JFP input is better than that of the S-branch using the skeleton sequence.
	After combining two branches, the final recognition accuracy is improved by 8.9\%  over the S-branch. When examined on a stronger GCN backbone, 2S-AGCN, the proposed JFP stream again shows its great value in improving the recognition accuracy. Specifically, our proposed P-branch achieves an X-sub accuracy of 88.1\%, which is comparable to 88.5\% obtained by the S-branch~(using joints and bones). 
	Combining the two complementary branches together is beneficial and leads to a much better recognition result (93.8\%) than that of each single branch alone. The accuracy gain brought by the JFP stream is around 5.3\% over the original S-branch. 
	To put this in context, merging the ``Joints” stream and the ``Bones” stream yields only an accuracy gain of less than 2\% in 2S-AGCN.
	
%-------------------------------------------------------
	\begin{table}[t]
		\centering
		\begin{center}
			\begin{tabular}{|l|c|c|}
				\hline
				Modality    			& 2S-AGCN~(\%) & ST-GCN~(\%)  \\
				\hline
				JAP		 			& 83.8 & 81.1  	   			  \\
				
				JDP      			& 85.9 & 82.9  	   			  \\
				
				JFP      			& 88.1 & \textbf{86.6}  	  \\
				
				Skeleton 			& \textbf{88.5} & 81.5 	   	  \\ % 使用原文结果
				\hline
				
				Skeleton + JAP       & 92.0 & 87.7  	   		  \\
				
				Skeleton + JDP       & 92.1 & 88.2  	   		  \\
				
				Skeleton + JFP       & \textbf{93.8} & \textbf{90.4} \\
				\hline
				
				Skeleton + JFP + JAP      & 93.9 & 91.2  	   	  \\
				
				Skeleton + JFP + JDP      & \textbf{94.1} & \textbf{91.8}\\ 
				\hline
			\end{tabular}
		\end{center}
		\caption{Comparisons of the accuracy obtained by different modalities and their combinations evaluated on the NTU RGB+D dataset using the X-sub protocol. When different modalities are combined, their respective weights are set equal.}
	    \vspace{-0.1in}
		\label{tab:different_sources}
	\end{table}
	%--------------------------------------------------------

\begin{figure*}
		\centering
		\includegraphics[width=\textwidth,height=2.3cm]{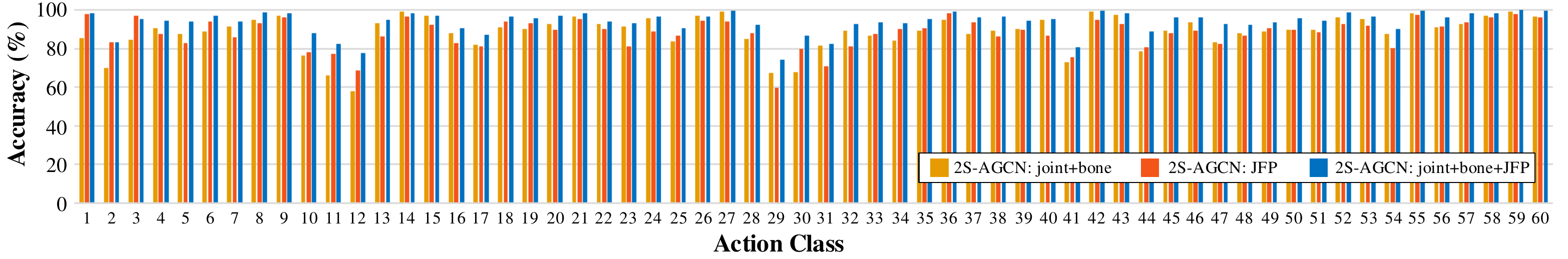}
		\caption{Class-by-class action recognition accuracy comparisons of the S-branch, P-branch and their combinations evaluated with the X-sub protocol on the NTU RGB+D dataset.}
		\label{fig:histogram}
	\end{figure*}

	\textbf{Evaluation of the JFP Gain over Each Action Class:} 
	To appreciate the accuracy gain by including the JFP stream in more depth, we compare the performances of different branches on every action category. 
	As shown in Figure~\ref{fig:histogram}, the performances of the single S-branch on 2S-AGCN~(``joint+bone") backbone encounter an accuracy bottleneck in the actions characterized primarily by local subtle movements, due to the limitations of the sparse skeleton sequence discussed in Section~\ref{sec:intro}. After merging S-branch and P-branch, these actions mainly characterized by local subtle movements, such as ``clapping"(\#10), ``reading"(\#11), ``writing"(\#12), ``play with phone/tablet"(\#29), and ``type on a keyboard"(\#30), have gained significant accuracy improvements, i.e., 11.3\%, 16.1\%, 19.8\%, 6.9\%, and 19.2\% for the experiments using the 2S-AGCN backbone, respectively. This per-action class recognition accuracy comparison shows clearly that the proposed JFP sequences effectively capture the local subtle motion information and help the network to make more accurate recognition.
	%------------------------------------------------------
	%------------------------------------------------------	
	
    %-------------------------------------------------------
    \begin{table*}[t]
    	\centering
    	\begin{center}
    		\begin{tabular}{|l|c|c|c|c|}
    			\hline
    			Method      & Parameters     & GPU Memory Consumption      & Runtime   & FLOPS  \\ 
    			\hline
    			ST-GCN      & 3.10M     &341MB      & 93ms    & 16.3G  \\
    			\hline
    			2S-AGCN     & 6.94M     &704MB      & 138ms      & 37.3G  \\
    			\hline
    			JOLO-GCN~( ST-GCN+P-branch)& 3.10+3.10M& 341+353MB & 93+22ms     & 16.3+3.9G  \\
    			\hline
    			JOLO-GCN~( 2S-AGCN+P-branch)& 6.94+3.48M& 704+468MB & 138+29ms & 37.3+4.5G  \\
    			\hline
    		\end{tabular}
    	\end{center}
    	\caption{Comparisons of the proposed JOLO-GCN and baselines in parameter, runtime, and resource consumption. The above statistics report the average values for one forward inference on a Tesla K80 GPU. FLOPS denotes floating-point operations per second.}
    	\label{tab:complexity}
    	\vspace{-0.05in}
    \end{table*}
    %--------------------------------------------------------

	\textbf{Comparison of Using Different Patch Modalities:} 
	In fact, it remains intriguing to find out whether other patch modalities can possibly be better options over JFP, such as JAP or JDP~(Joint-centered Depth Patches). 
	To this end, we conduct a series of experiments to evaluate different patch modalities, and their combinations with the input skeleton stream. The results are reported in Table~\ref{tab:different_sources}.
	
	A few key observations can be found from Table~\ref{tab:different_sources}. First, when a single modality is used for GCN-based action recognition, the input skeletons or the proposed JFP stream tend to give better accuracies than JAP or JDP, as they capture motion dynamics more important and direct. Second, if one additional patch modality is added to complement the original skeleton stream, JFP is again the best choice over JAP and JDP, leading to the recognition accuracy of 93.83\% on the 2S-AGCN backbone. Such a combination actually corresponds to our proposed method evaluated earlier in Table~\ref{tab:NTU+KS}. Third, it can be found that including one more extra patch modality besides JFP at the cost of increased network complexity is not necessary, because the accuracy gain is quite marginal, e.g., 94.1\% (Skeleton + JFP + JDP) versus 93.8\% (Skeleton + JFP). Also, depth maps are not always available as the input to extract the JDP stream from.

    \textbf{Runtime Analysis:}
    The following discussions are conducted using the NTU RGB+D dataset. In our proposed method, the JFP pre-processing includes loading full-HD video, temporal downsampling, joint-centered cropping of image patches (JAP), and TV-L1 optical flow estimation for computing JFP. The original videos of NTU RGB+D have 80 frames on average, and we downsampled them by a temporal factor of 2. Our JFP pre-processing takes about 1.5 seconds for a video averagely on a laptop with a Intel i5-7300HQ CPU without using GPUs. Our method using the 2S-AGCN backbone~(three streams: ``Joints"+``Bones"+``JFP") takes about 167 ms for a full prediction using a Tesla K80 GPU.
    
    \textbf{Model Complexity Analysis:}
    Table~\ref{tab:complexity} shows the comparison of JOLO-GCN and baseline methods~(ST-GCN and 2S-AGCN) in model complexity. Compared with the baseline methods, the computational complexity introduced by the P-branch is relatively low. The difference of the joints numbers and frames numbers between skeleton sequences~(25 joints, 300 frames) and JFP sequence~(14 joints, 64 frames) makes P-branch fast computationally compared with both baselines. 

	\section{Conclusion}
	In this paper, we proposed a novel approach of representing the visual information surrounding each skeletal joint as Joint-aligned optical Flow Patches~(JFP), effectively capturing the useful local subtle body motion cues for skeleton-based action recognition. 
	The derived JFP sequence has the advantage of a compact representation and inherits a kinetically meaningful structure from the human pose skeleton. Based on the proposed JOLO-GCN framework, we jointly exploit local subtle motion cues from the JFP sequence and global motion cues from the skeleton sequence for action recognition. 
	The proposed method obtains state-of-the-art results on the three large-scale action recognition datasets. 
	Our experiments further show that when applied on two different GCN-based backbones~(ST-GCN~\cite{yan2018spatial} and 2S-AGCN~\cite{shi2019two}), 
	the proposed method improves both of them by large performance margins. This validates the generalization ability of applying our scheme on different light-weight single-modal skeleton-based networks.

	\section*{Acknowledgement}
	The work was supported in part by the Key Area R\&D Program of Guangdong Province with grant No.~2018B030338001, by the National Key R\&D Program of China with grant No.~2018YFB1800800, by Natural Science Foundation of China with grant NSFC-61902334 and NSFC-61629101, by Guangdong Zhujiang Project No.~2017ZT07X152, and by Shenzhen Key Lab Fund No.~ZDSYS201707251409055, by the Program for Guangdong Introducing Innovative and Enterpreneurial Teams (Grant No.:~2017ZT07X183), the National Natural Science Foundation of China (Grant No.:~61771201), and the Guangdong R\&D key project of China (Grant No.~2019B010155001).
	
{\small
\bibliographystyle{ieee_fullname}
\bibliography{egbib}
}

\end{document}